\documentclass{elsarticle}
\usepackage{multirow}
\usepackage{xcolor}

\usepackage{verbatim}

\usepackage{inputenc}

\usepackage{graphicx}

\usepackage{amssymb}

\usepackage[modulo]{lineno}
\modulolinenumbers[2]

\usepackage[hyphens]{url}

\usepackage[bookmarks, colorlinks, breaklinks,unicode=true]{hyperref}

\usepackage{algorithm}
\usepackage[noend]{algpseudocode}

\hypersetup{linkcolor=blue,citecolor=blue,filecolor=black}

\usepackage[a4paper, total={6in, 8in}]{geometry}

\usepackage{outlines}

\usepackage{textcomp}
\usepackage{gensymb}

\usepackage{enumitem}

\usepackage{amssymb}

\usepackage{pifont}

\usepackage{subfigure}

\usepackage[numbers]{natbib}
\newlist{todo}{itemize}{2}
\setlist[todo]{label=$\square$}
\usepackage[toc, section=section, acronym]{glossaries}
\usepackage[toc, section=section]{glossaries}

\usepackage{glossary-mcols}

\setacronymstyle{long-short}

\setglossarystyle{mcolindex}

\newacronym{GFS}{GFS}{Global Forecast System}
\newacronym{GEFS}{GEFS}{Global Ensemble Forecast System}
\newacronym{ECMWF}{ECMWF}{European Center for Medium Weather Forecasting}
\newacronym{AnEn}{AnEn}{Analog Ensemble}
\newacronym{PAnEn}{PAnEn}{Parallel Analog Ensemble}
\newacronym{NWP}{NWP}{Numerical Weather Prediction}
\newacronym{PAN}{PAN}{Persistence Analog}
\newacronym{GCM}{GCM}{General Circulation Model}
\newacronym{NAM}{NAM}{North American Mesoscale Model}
\newacronym{WRF}{WRF}{Weather Research and Forecasting}
\newacronym{SSE}{SSE}{Search Space Extension}
\newacronym{EA}{EA}{Evolutionary Analog}
\newacronym{GA}{GA}{Genetic Algorithm}
\newacronym{KF}{KF}{Kalman Filter}
\newacronym{CRM}{CRM}{Cloud Resolving Models}
\newacronym{HRRR}{HRRR}{High-Resolution Rapid Refresh}
\newacronym{SAM}{SAM}{System Advisor Model}
\newacronym{HF}{HF}{Heuristic Filter}
\newacronym{LOO}{LOO}{Leave-One-Out}
\newacronym{SS}{SS}{Schaake Shuffle}
\newacronym{SSI}{SSI}{Spectral Statistical Interpolation}
\newacronym{GSI}{GSI}{Gridpoint Statistical Interpolation}
\newacronym{UMAP}{UMAP}{Uniform Manifold Approximation and Projection}
\newacronym{SOM}{SOM}{Self-Organizing Map}

\newacronym{NCEP}{NCEP}{National Centers for Environmental Prediction}
\newacronym{WU}{WU}{Weather Underground}
\newacronym{PWS}{PWS}{Private Weather Station}
\newacronym{NOAA}{NOAA}{National Oceanic and Atmospheric Agency}
\newacronym{WHO}{WHO}{World Health Organization}
\newacronym{ATSDR}{ATSDR}{Agency for Toxic Substances and Diseases Registry}
\newacronym{CMIP}{CMIP}{Coupled Model Intercomparison Project}
\newacronym{IPCC}{IPCC}{International Panel on Climate Change}
\newacronym{ASOS}{ASOS}{Automated Surface Observing System}
\newacronym{NCAR}{NCAR}{National Center for Atmospheric Research}
\newacronym{GISS}{GISS}{Goddard Institute for Space Studies}
\newacronym{RCP}{RCP}{Representative Concentration Pathway}
\newacronym{WRCP}{WRCP}{World Climate Research Programme}

\newacronym{SVI}{SVI}{Social Vulnerability Index}
\newacronym{ME}{ME}{Mean Error}
\newacronym{RMSE}{RMSE}{Root Mean Square Error}
\newacronym{CRMSE}{CRMSE}{Centered Root Mean Square Error}
\newacronym{RC}{RC}{Rank Correlation}
\newacronym{MSE}{MSE}{Mean Square Error}
\newacronym{OMT}{OMT}{Over Max Time}
\newacronym{CRPS}{CRPS}{Continuous Rank Probability Score}
\newacronym{RPS}{RPS}{Rank Probability Score}
\newacronym{MAE}{MAE}{Mean Absolute Error}
\newacronym{Brier}{Brier}{Brier Score}

\newacronym{PV}{PV}{Photovoltaic}
\newacronym{AVS}{AVS}{Agrivoltaic System}
\newacronym{LBR}{LBR}{Land-Based Renewables}
\newacronym{USSE}{USSE}{Utility-Scale Solar Energy}

\newacronym{VGI}{VGI}{Volunteered Geographic Information}
\newacronym{UHI}{UHI}{Urban Heat Island}
\newacronym{CONUS}{CONUS}{Continental United States}

\newacronym{IDW}{IDW}{Inverse Distance Weighted}
\newacronym{PDF}{PDF}{Probability Distribution Function}
\newacronym{CDF}{CDF}{Cumulative Distribution Function}

\newacronym{API}{API}{Application Programming Interface}
\newacronym{PEF}{PEF}{Parallel Ensemble Program}
\newacronym{CLI}{CLI}{Command Line Interface}
\newacronym{RAM}{RAM}{Random Access Memory}
\newacronym{UML}{UML}{Unified Modeling Language}
\newacronym{TAU}{TAU}{Tuning and Analysis Utilities}
\newacronym{HPC}{HPC}{High-Performance Computing}
\newacronym{ANN}{ANN}{Artifitial Neural Network}
\newacronym{ML}{ML}{Machine Learning}
\newacronym{LSTM}{LSTM}{Long Short-Term Memory}
\newacronym{FNN}{FNN}{Feedforward Neural Network}
\newacronym{GNN}{GNN}{Graph Neural Network}

\newacronym{FLT}{FLT}{Forecast Lead Time}
\newacronym{ITCZ}{ITCZ}{Intertropical Convergence Zone}
\newacronym{ENSO}{ENSO}{El Ni\~no Southern Oscillation}
\newacronym{PDO}{PDO}{Pacific Decadal Oscillation}
\newacronym{MOS}{MOS}{Model Output Statistics}
\newacronym{PNA}{PNA}{Pacific/North American}
\newacronym{GHG}{GHG}{Greenhouse gas}

\newcommand*{\Fig}[1]{Figure~\ref{#1}}

\newcommand*{\Table}[1]{Table~\ref{#1}}

\title{Using Long Short-Term Memory (LSTM) and Internet of Things (IoT) for localized surface temperature forecasting in an urban environment}

\author[GEOG-ICDS]{Manzhu Yu}

\author[GEOG-ICDS]{Fangcao Xu}

\author[GEOG-ICDS]{Weiming Hu}

\author[GEOG-ICDS]{Jian Sun}

\author[GEOG-ICDS,NCAR]{Guido Cervone}

\address[GEOG-ICDS]{Department of Geography and Institute for Computational and Data Science, The Pennsylvania State University, University Park, PA, USA}

\address[NCAR]{Research Application Laboratory (RAL), National Center for Atmospheric Research (NCAR), Boulder, CO, USA}

\begin{document}

\begin{abstract}
The rising temperature is one of the key indicators of a warming climate, and it can cause extensive stress to biological systems as well built structures. Due to the heat island effect, it is most severe in urban environments compared to other landscapes due to the decrease in vegetation associated with a dense human built environment.  It is essential to adequately monitor the local temperature dynamics to mitigate risks associated with increasing temperatures, which can include short term strategy to protect people and animals, to long term strategy to how to build new structure and cope with extreme events.  Observed temperature is also a very important input for atmospheric models, and accurate data can lead to better future forecasts.

Ambient temperature collected at ground level can have a higher variability when compared from regional weather forecasts, which fail to capture the local dynamics. There remains a clear need for an accurate air temperature prediction at the sub-urban scale at high temporal and spatial resolution. This research proposed a framework based on Long Short-Term Memory (LSTM) deep learning network to generate day-ahead hourly temperature forecast with high spatial resolution.  A case study is shown which uses historical in-situ observations and Internet of Things (IoT) observations for New York City, USA. By leveraging the historical air temperature data from in-situ observations, the LSTM model can be exposed to more historical patterns that might not be present in the IoT observations. Meanwhile, by using IoT observations, the spatial resolution of air temperature predictions is significantly improved.
\end{abstract}


\maketitle

\section{Introduction}

One of the significant aspects of climate change is globally rising temperature. According to the NOAA 2020 Global Climate Summary, the land and ocean surface temperature of August 2020 was 0.94°C (1.69°F) above average and ranked the second highest August temperature since 1880 \cite{dunn2020global}. Rising temperature causes extreme heat phenomena, such as urban heat islands, producing life-threatening conditions, overheating rivers, plants, and wildlife. The urban heat island (UHI) is a phenomenon that urban areas have higher temperatures (1-7\degree F) than the surrounding rural areas \cite{oke1973city}. Apart from the overall rising temperature, urban heat islands can be caused by reduced natural landscapes in urban areas, urban material properties reflecting less solar energy, urban geometries hindering wind flow, and heat generated from human activities. Over 55\% of the world's population lives in urban areas, which is predicted to reach 68\% by 2050 \cite{UN2019}. Therefore, increasing risks of heat-related deaths and illnesses and increasing demands of power exist in urban areas. 

It is essential to adequately monitor the local temperature dynamics to mitigate risks associated with increasing global temperatures. For that purpose, having good spatiotemporal coverage of temperature measurements is one necessity. Regional weather forecasts provide a spatiotemporally continuous estimate of weather conditions, but such estimates are still limited in their spatial resolution, especially for personal or street-level uses \cite{pelta2017spatiotemporal}. Localized weather can be quite different from the regional weather forecast. Localized heat forecast can help identify the regions which are prone to overheating and target warnings to citizens on potential heatwaves and provide aid to residents in time \cite{shi2018modelling}. There is an important need for accurate hourly air temperature measurements at very high spatial resolution in urban environments. 

Although high spatial-resolution weather simulation models can produce local forecasts, the accuracy of the predictions and future mitigation decisions are still heavily influenced by the availability of observations -- the ground truth. These mitigation suggestions can be salting icy roads, turning on public water sprays, or providing shelters to the public in extreme weather situations \cite{chapman2018high}. The verification of model forecasts also requires high-resolution observations. However, traditional monitoring infrastructures cannot provide such information due to the limited number of discrete stations installed. The increasing availability of Internet of Things (IoT) can provide an excellent complement to the traditional in-situ observations regarding local uncertainty. For example, the Array of Things network in Chicago has embedded around 150 sensors in the city to monitor urban climate in a community level \cite{catlett2017array}. The surface temperature has also been estimated using smartphones' battery temperature in a crowdsourced way where proper quality control is conducted \citep{overeem2013crowdsourcing,muller2015crowdsourcing}. The Fifth Generation (5G) of mobile technologies and its potential impacts on IoT will bring enormous benefits to localized weather observation with higher data transmission speed and more connected networks \cite{alawe2018improving}. 

This research proposed a framework by integrating long-term historical in-situ observations and IoT observations together to train a Long Short-Term Memory (LSTM) network for air temperature prediction within the city of New York. By leveraging the historical air temperature data from in-situ observations, the LSTM model can be exposed to more historical patterns that might not be present in the IoT observations. Meanwhile, by using IoT observations, the spatial resolution of air temperature predictions is significantly improved. 

The paper is organized as follows. The related works are reviewed and discussed in Section 2. In-situ and IoT datasets used in this study are introduced in Section 3. The LSTM model adopted in this study is described in Section 4. Experiment results are reported in Section 5, and extreme cases are demonstrated in Section 6, followed by the conclusions and discussion in Section 7.

\section{Related works}

\subsection{IoT for urban temperature monitoring}

An increasing number of cities are implementing urban meteorological monitoring projects of differing size and scales as part of "smart city" initiatives and scientific research projects, including the Birmingham urban climate laboratory \cite{chapman2015birmingham}, the Safe Community Alert Network at Montgomery County \cite{benson2015scale}, the Array of Things network in Chicago \cite{catlett2017array}, and the Smart Santander in Spain \cite{sotres2017practical}. These initiatives and research projects have provided unprecedented new opportunities for high-resolution monitoring of the urban climate. Moreover, these monitoring helps the city become smarter by controlling energy demand and reducing transport network disruption. 

For temperature studies, increasing the spatiotemporal resolution of urban meteorological monitoring becomes even more crucial. Street-level air temperature has a tremendous spatiotemporal variability that impacts the vulnerable population in different ways. For example, the Array of Things network in Chicago installed $\sim$150 stationary devices ("nodes") in Chicago, typically at street intersections, to monitor the city's climate, noise level, and air quality \cite{catlett2017array}. However, these "nodes" help increase the density of monitoring only to a certain extent, where each "node" covers a community instead of a street. The Smart Santander project is now embedding the city with more than 12,500 sensors \cite{sotres2017practical}. A portion of sensors is mounted on stationary objects, such as trash containers, streetlights, parking spaces. In contrast, the other sensors are mounted on vehicles such as police cars and taxicabs that monitor air pollution and traffic conditions. The data collected from the larger number of sensors leads to improvements in urban weather monitoring and a better grasp of urban issues. 

Utilizing their high spatiotemporal resolution, researchers have explored IoT infrastructures to monitor urban climate and assist in various urban issues. Chapman et al. \cite{chapman2016using} used the IoT network to measure rail moisture and leaf-fall contamination to achieve a low-cost, real-time, and high spatiotemporal resolution rail monitoring system. Chapman and Bell \cite{chapman2018high} demonstrated a use case utilizing IoT sensors to obtain high-resolution temperature observations for winter road maintenance. These observations are used in route-based forecasting models to decide which road segments need salting treatments in snow or icy situations. Ferranti et al. \cite{ferranti2016heat} utilized an IoT network to monitor the temperature rise along railways and analyze the relationship between railway failure and gradual rise in temperature during the early or mid-summer season. This relationship could be useful in heat risk management and potentially reduce disruptions and delays in the railway services. Kraemer et al. \cite{kraemer2020operationalizing} utilized an IoT system with solar power and weather forecasts to predict solar power energy. They selected relevant features from weather forecasts and trained machine learning models that generate predictions with 20\% better accuracy than the current state of the art prediction. The solar energy predictions can be used for effective energy budget planning. Using IoT hardware, software, and communication technologies, Shapsough et al. \cite{shapsough2018using} developed a cost-effective system for the stakeholders to monitor and efficiently control the large-scale solar photovoltaic systems and evaluate the effects of environmental factors to the systems.

In this research, we use the data collected by a vehicle-based IoT network which collects air temperature information along major roads in and around big cities in the US. The collected information is then preprocessed to eliminate outliers and noises and aggregated hourly into $\sim$150m $\times$ 150m grids. The high spatial and moderately high temporal resolution provides us a great opportunity to monitor air temperature in a sub-urban scale. 

\subsection{Machine learning temperature prediction}

Air temperature prediction is one of the most important aspects of climate study. An accuracy temperature prediction can provide crucial guidance for the decision-making process to address environmental, ecological, or industrial problems. Machine learning techniques have been used in air temperature predictions based on the time series of historical air temperature and possibly other input predictors, such as humidity, wind speed and directions, and surface pressure. Example machine learning methods include Support Vector Machines (SVM), Artificial Neural Networks (ANN), and more recently Convolutional Neural Networks (CNN) and LSTM Recurrent Neural Networks (RNN). Some works have explored the capability of using machine learning methods to predict global temperature under climate change for the future decadal or longer time scales \citep{fildes2011validation,abubakar2016utilising,hassani2018predicting}. Most of these works demonstrated the impacts of CO2 emissions on global temperature rising and compared the global temperature predictions generated by machine learning methods with the Intergovernmental Panel on Climate Change (IPCC) scenarios \cite{fildes2011validation}. However, global temperature models do not provide local or regional forecasts with fine-scaled air temperature variability. 

Other works have been integrating observations from weather stations into the machine learning models for regional or local air temperature forecasting on an hourly or daily basis. For example, Smith et al. \cite{smith2009artificial} used a Ward-style ANN with historical 24-hour air temperature, wind speed, precipitation, relative humidity, and solar radiation to predict the air temperature at one or multiple future hours for the state of Georgia. Ward-style ANNs are single-layer feedforward neural networks that utilize backpropagation and activation functions to optimize weights and biases. Results showed an increasing error when predicting a longer time period in the future, with the Mean Absolute Error (MAE) ranging from 0.516 °C at the one-hour horizon to 1.873 °C at the twelve-hour horizon. Focusing on the same region, Chevalier et al. \cite{chevalier2011support} compared the support vector regression (SVR) and the single-layer ANN in the capability of predicting sudden changes in air temperature and observed different capabilities of the two models in predicting year-round and winter-only dataset. Results showed that SVR predicted more accurately for the year-round dataset, whereas ANN generally outperformed SVR using the winter-only dataset. Adding more hidden layers, Hossain et al. \cite{hossain2015forecasting} trained a three-layer feedforward neural network with historical 24-hour air temperature from weather stations, barometric pressure, humidity, and wind speed to predict the air temperature at a certain future hour. More recently, Hewage et al. \cite{hewage2020deep} trained and compared two deep learning models (LSTM and Convolution RNN) with surface temperature, pressure, wind, precipitation, humidity, snow, and soil temperature that are generated from numerical weather prediction models. Both models are used to predict air temperature for a certain future hour (one-step prediction), and both models are composed of five hidden layers. 

Most of the works mentioned above focus on one-step prediction instead of a multi-step prediction. These works have been using data collected from weather stations or numerical simulations that generally have a coarse spatial resolution and cannot provide street-level air temperature variability. There have been various researches using IoT networks and machine learning techniques for air quality prediction \citep{kok2017deep,jin2019high}, agriculture frost prediction \cite{brun2016demo}, or building heating and cooling demand prediction \cite{luo2019development}. However, to the authors’ knowledge, this is the first research integrating traditional in-situ observations and IoT observations to improve the multi-step predicting capability of deep learning techniques.

\section{Study area and data }

In this study, we focus on exploring the air temperature in New York City (NYC). Table \ref{table1} lists the sources, spatiotemporal resolutions, and statistical information of the datasets used in this study. The IoT data is from the GeoTab data platform, and the data availability ranges from Apr 29, 2019 to May 1, 2020. We also downloaded the air temperature from discrete weather stations in Weather Underground for historical weather information.

\begin{table}[ht!]
\caption{Summary of data collections}
\label{table1}
\begin{tabular}{l|l|l}
\hline
\textit{\textbf{Data type}} & \textit{Internet of Things (IoT)} & \textit{Weather stations} \\ \hline
\textbf{Data source} & \begin{tabular}[c]{@{}l@{}}GeoTab\\ (https://data.geotab.com/)\end{tabular} & Weather Underground \\ \hline
\textbf{\begin{tabular}[c]{@{}l@{}}Spatial and temporal\\ resolutions\end{tabular}} & \begin{tabular}[c]{@{}l@{}}153m x 153m grids along the\\ major roads for every 60 minutes\end{tabular} & \begin{tabular}[c]{@{}l@{}}Fixed sensor location for every\\ 60 mins\end{tabular} \\ \hline
\textbf{Time range} & May 1, 2019 to Apr 30, 2020 & Jan 1, 2015 to Apr 30, 2019 \\ \hline
\textbf{Number of stations} & 36970 & 130 \\ \hline
\textbf{Value range} & {[}-12.20, 48.00{]} \textbackslash{}degree C & {[}-22.55, 36.94{]} \textbackslash{}degree C \\ \hline
\textbf{Missing data ratio range} & {[}5.25\%, 100\%{]} & 0.076\% \\ \hline
\end{tabular}
\end{table}

\subsection{GeoTab}
GeoTab data includes ambient air temperature collected from sensors mounted on vehicles. This research uses the data where anomalous data and outliers have already been removed.  The measurements are aggregated to the 7-character geohash level (153m x 153m) every 60 minutes. Geotab tracks over 900,000 vehicles and generates temperature data from over 250,000 vehicles per hour throughout North America. This allows Geotab to accumulate millions of temperature data points per hour in near real-time and generate a comprehensive, continuously updating map of road temperature. 

\subsection{Weather Underground}
To build a reliable temperature prediction model, a long-term historical air temperature dataset pertaining to climate patterns is required. The availability of GeoTab, however, lasts only one year, indicating it is not adequate to build the prediction model with GeoTab alone. Contrary to GeoTab, data collected from weather stations are generally archived for a long period of time. The dataset we used in this study is the Weather Underground (WU), and we downloaded the hourly air temperature from WU for 2015-2020. WU is a network of weather stations that combines the authoritative observing systems and person weather stations. The authoritative observing systems include the Automated Surface Observation System (ASOS) stations located at airports throughout the country and the Meteorological Assimilation Data Ingest System (MADIS) managed by the National Oceanic and Atmospheric Administration (NOAA). The Federal Aviation Administration maintains ASOS stations, and observations are updated hourly, or more frequently when adverse weather affecting aviation occurs (low visibility, precipitation, etc.). The personal weather stations (PWS) are contributed by volunteers who purchase and install weather sensors in and around their houses or workplaces. These PWS stations are put through strict quality controls, and observations are updated as often as every 2.5 seconds. There are 130 stations within the bounding box of our study area, and the missing data rate for all these stations are mostly less than 0.005. We directly filled the data with NA for each station by their missing time points.

\subsection{Missing data handling}

\begin{figure}[H]
\centering
\includegraphics[width=1.0\textwidth]{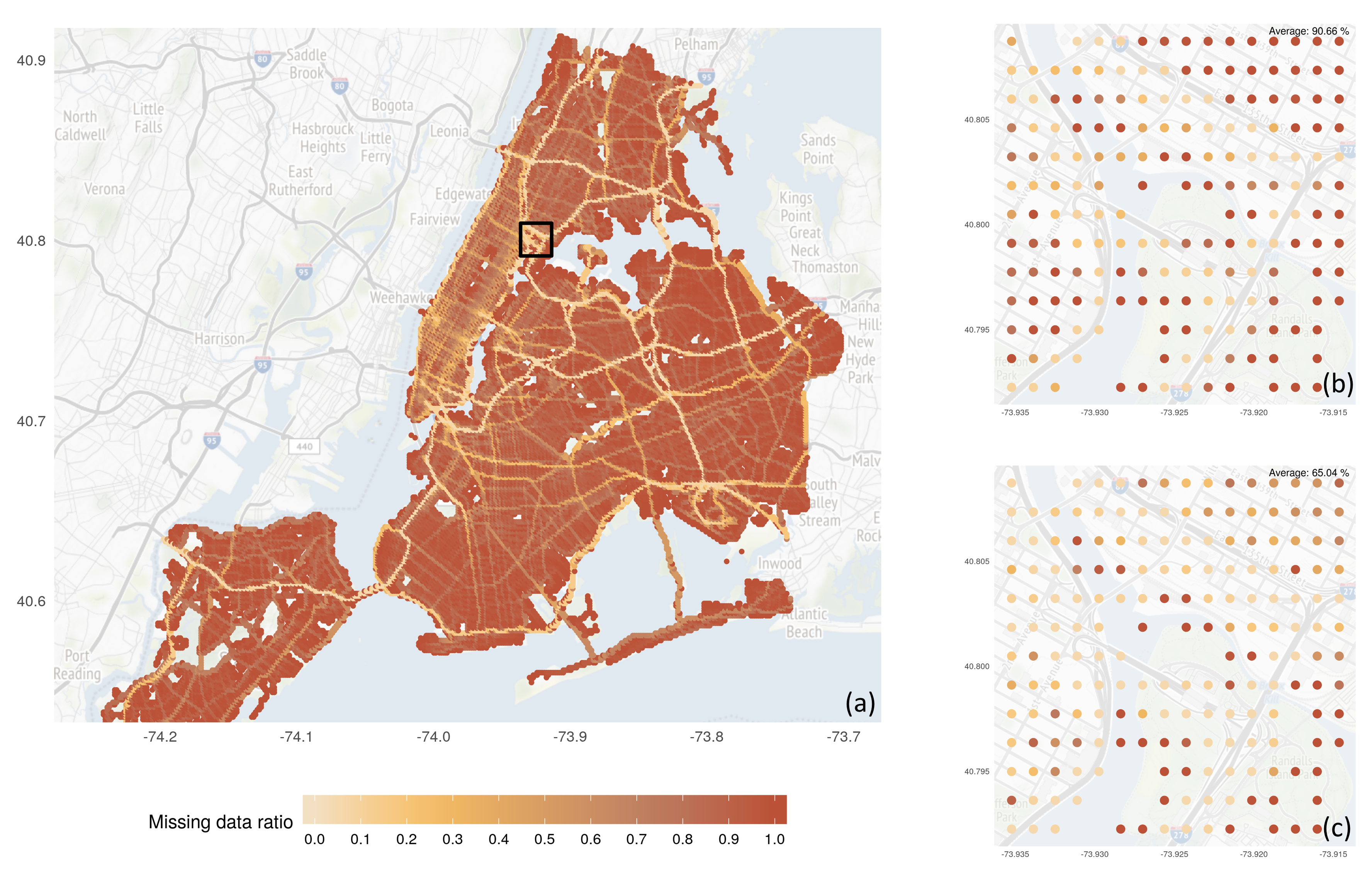}
\caption{
(a) Overall missing data ratio for each GeoTab grid, (b) Spatial distribution of missing data ratio for the same subregion at 4:00AM local time (having the maximum overall missing ratio), and (c) Spatial distribution of missing data ratio for the same subregion at 8:00AM local time (having the minimum overall missing ratio).
}
\label{fig1}
\end{figure}

Despite the high spatiotemporal resolution, an obvious disadvantage of vehicle-based measurement is the missing data issue due to a small (or even zero) number of vehicles passing the same location within that one hour. The aggregated air temperature data for that spatiotemporal grid have inferior quality or have missing data. Here we demonstrate the missing data rate of the GeoTab data. Figure \ref{fig1}a shows the spatial distribution of all 36970 GeoTab grids. Note that a particular grid may not have data since there might not be enough vehicles passing that grid in a specific hour. Thus, each grid has a different ratio of missing data, and similarly, each hour has a different spatial distribution based on the data availability. 

The missing data ratios for some GeoTab stations are relatively high compared to the WU dataset. To deal with this problem, we processed the GeoTab data in two steps: 1) select GeoTab grids along the major roads with a missing data ratio less than a certain threshold (i.e., under 5.5\%, 10\%, 20\%, …, 50\%), and 2) linearly interpolate the data of each grid in time, and then find nearest 20 stations for average. We interpolated the missing data linearly for each grid based on temporal dependencies. These grids are distributed along major roads, which is reasonable since it is a vehicle-based data collection. There are more vehicles on major roads than the ones on other secondary roads. This data characteristic makes this study more specific, focusing on the air temperature prediction along major roads of NYC. 

\section{Methods}

\subsection{LSTM}

Recurrent neural networks (RNNs) have been used to learn the sequential patterns in time series data. Taking the current and the previous status, a hidden state at a time step t of an RNN can take the memory forward to predict the next time step (t+1). LSTM is a typical kind of RNNs and can learn a more extended period than a simple RNN \cite{hochreiter1996lstm}. The hidden state of LSTM can be controlled at the gates to reduce the vanishing gradient and the exploding gradient problems that are usually suffered by RNNs \cite{gers1999learning,graves2008novel}. 

\begin{figure}[ht!]
\centering
\includegraphics[width=1.0\textwidth]{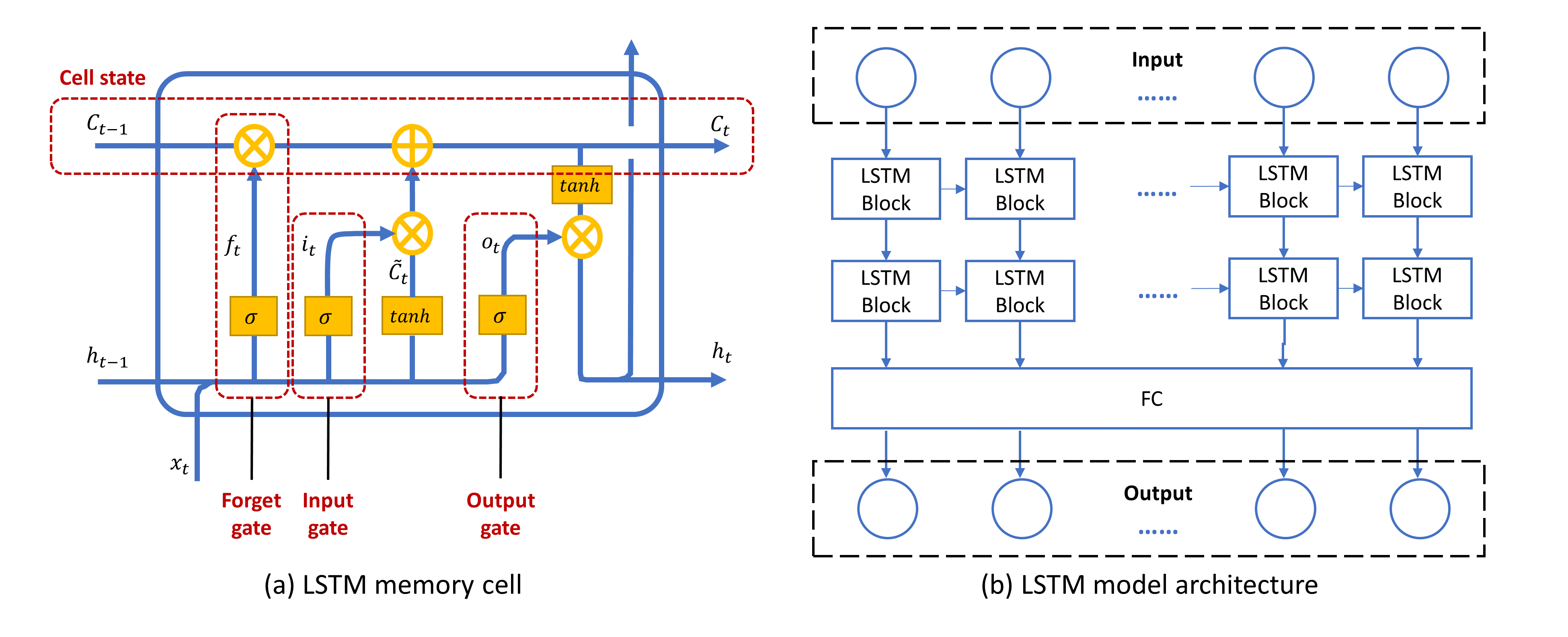}
\caption{
The architecture of a LSTM cell.
}
\label{fig2}
\end{figure}

The key to LSTM is the cell state, and adding or removing information to or from the cell state is achieved by gates, which is composed of a sigmoid neural net layer and a pointwise multiplication operation. The sigmoid layer's output is between 0 and 1, with 0 indicating letting no information passing through, and 1 indicating all information. An LSTM cell contains three gates: input gate, output gate, and forget gate (Figure \ref{fig2}a). Within each cell, the first step is to select the cell state from the previous time step and retain part of the information into the current time step using the forget gate. The forget gate is described as Equation 1, where the hidden state of the previous time step ($h_{t-1}$) and the value of the current time step ($x_t$) are taken into account in the sigmoid function $\sigma{(\cdot{})}$. The second step is to control the inward information into the cell using the input gate. This process is conducted using a sigmoid layer (Equation 2) that determines which values of the cell state to update and a tanh layer (Equation 3) that creates intermediate values ($\tilde{C_t}$) to update the cell state (Equation 4). The last step is to control the outward information from the cell. This process is achieved by a sigmoid layer (Equation 5) that determines which values of the cell state to output and a tanh layer that standardizes the values of the cell state. The sigmoid layer and the tanh layer are then multiplied to calculate the current hidden state (Equation 6).  

\begin{equation}
    f_t=\sigma(W_f\cdot[h_{t-1},x_t]  + b_f )
\end{equation}

\begin{equation}
    i_t= \sigma(W_i\cdot[h_{t-1},x_t ]  + b_i )  
\end{equation}

\begin{equation}
    \tilde{C_t}= tanh(W_C\cdot[h_{t-1},x_t ]  + b_C)    
\end{equation}

\begin{equation}
    C_t=f_t*C_{t-1}+i_t*\tilde{C_t}
\end{equation}

\begin{equation}
    o_t= \sigma(W_o\cdot[h_{t-1},x_t ]  + b_o )
\end{equation}

\begin{equation}
    h_t=o_t*tanh(C_t)
\end{equation}

In Equations 1-6, the sigmoid functions are calculated as \(\sigma(x)=  1/(1+e^{-x} )\)
, the tangent function is calculated as \(tanh(x)=(e^x-e^{-x})/(e^x+e^{-x}))\), $W_f$, $W_i$, $W_C$, $W_o$ are the weight matrices, $b_f$, $b_i$, $b_C$, $b_o$ are the bias vectors, $x_t$ is the current input, $h_{t-1}$ and $h_t$ are the hidden states of the previous time step and the current time step. 

LSTM models are a natural fit for our problem due to the following two reasons. First, LSTM is capable of handling long sequential data processing because the design of gates allows intact memory propagation, shown as the state passing, which avoids, to some extent, the gradient vanishing and exploding issues. Second, comparing to conventional RNN, LSTM is relatively insensitive to the ``gap'' length, i.e., the 'interval' between two adjacent cells. Temperature data have a similar characteristic because extreme weather may break the internal pattern existed in temperature. The gates design is extremely useful to eliminate the outlier data when finding the interior pattern of the time-series data. For example, the forget gate has the capacity to fully block the cell state memory passed from the previous time step. 

The LSTM architecture used in this study is a conventional LSTM neural network for temperature prediction, which consists of N number of LSTM layers and one fully connected (FC) layer (Figure \ref{fig2}b). The input data is feature vector of surface temperature observed for the past time series. The input data is fed into the stacked N layers of LSTM cells, where N is a tunable hyperparameter. The output of LSTM cells can be stacked into a matrix as input of the next layer. An LSTM layer is comprised of a set of M hidden nodes, where M is another tunable hyperparameter. When a single sequence of length $sl_in$ is passed into the network, each individual element of the sequence is passed through each and every hidden node. Each hidden node gives a single output for each input it sees, which results in an overall output from the hidden layer of shape $(sl_in,M)$. After the set of LSTM layers, a FC layer is added to the network for final output. The input size of the final FC layer is equal to the number of hidden nodes in the LSTM layer that precedes it. The output of this final FC layer is dependent on the output sequence length, $sl_out$, that the model will predict. For the multi-step temperature prediction, we used the Mean Squared Error (MSE) loss and the Adam optimizer. 

\subsection{Performance evaluation}

To evaluate the model performance, we used the Root Mean Squared Error (RMSE) and bias error (Bias). RMSE and Bias are on the same scale as the data and are calculated as:
\begin{equation}
    RMSE =\sqrt{\frac{1}{n} \sum_{i=1}^{n}(\tilde{y_i}-y_i)^{2}}
\end{equation}
\begin{equation}
    Bias = \tilde{y_i}-y_i
\end{equation}
where $\tilde{y_i}$ is the prediction and $y_i$ is the ground truth for data sample $i$. Note that each data sample $i$ contains a vector with a length of 24. RMSE and Bias are calculated for each station and each test sample, and the 24-hour predictions are evaluated as a single entity. Specifically, Bias is calculated as the mean bias for the 24-hour predictions. 

\subsection{Training procedure}
To demonstrate the capability of the LSTM architecture, the model is trained using 90\% and validated using 10\% GeoTab dataset and WU dataset. The model is tested using 27 randomly selected time series of continuous 72 (previous 48 + targeting 24) for GeoTab stations with up to 5.5\% missing data ratio. The 27 different days within the time range of the GeoTab observations are between 2019-05-01 and 2020-04-30 under different weather conditions. In addition, we ensure that the training and testing data are across all available months so that there is no bias or difference in the difficulty of predicting the test data. All experiments conducted are using the same testing days for a fair comparison. The GeoTab stations used in the testing data are the ones with up to 5.5\% missing data, where values are interpolated the least. 

\subsection{Hyperparameter tuning}

The effect of different combinations of the numbers of LSTM layers, hidden nodes and other hyperparameters on the prediction accuracy is investigated by changing the LSTM layers from 1 to 3, hidden nodes from 24 to 72, and learning rate from 0.05 to 0.005. While tuning the hyperparameters, we observed the following characteristics:
\begin{itemize}
    \item When the learning rate decreases from 0.05 to 0.005, the training process takes longer but persists for more training epochs before the model gets overfitting. The model trained with a learning rate of 0.05 has the largest RMSE, and decreasing the learning rate to 0.01 results in a 3\% decrease of RMSE, while decreasing the learning rate to 0.005 results in a 10\% decrease of RMSE.   
    \item The LSTM architecture with three layers outperforms the one with two layers by 20\%, given the hidden node size is set to be 24.
    \item Increasing the hidden size from 24 to 48 using the two-layer LSTM architecture can achieve similar accuracy to three layers but 24 nodes, but the the two-layer architecture is more efficient than the three-layer one.
    \item Continuously increasing the number of hidden nodes from 48 to 72 and the number of hidden layers from 2 to 3 does not significantly improve the accuracy. Still, the training time is three to five times longer than the one with 48 nodes and two layers. 
\end{itemize}

Based on the characteristics discussed above, the hyperparameters used in the experiments are listed below. The number of hidden layers is selected as 2. The number of hidden nodes is selected as 48, with a learning rate of 0.005 to balance model accuracy and training efficiency.

The number of epochs is determined by initiating a large enough best loss and updating it at each epoch and applying the early stopping technique to avoid overfitting.  Comparing each epoch's running loss with the best loss recorded, if the running loss is ten times larger than the best loss, the training process is terminated.  We also found the number of batches between 5 and 10 for each station can avoid overfitting too fast in very few epochs. Thus, the batch size is set as 500 and 5000 for GeoTab station and WU station, respectively. These batch sizes correspond proportionally to the number of training samples for each GeoTab, and WU station is around 4000 and 40000. 



\section{Experiment results}
To demonstrate the predicting capability of the proposed approach, we compared the model performance with other commonly used time series forecasting methods. To understand the impact of missing data on our proposed approach's predictability, we also conducted a sensitivity experiment by changing the missing data ratio from 5.5\% to 50\%. To understand the impact of adding historical WU data to the training, a comparison experiment was conducted to examine whether and how the historical WU data can improve the local weather predictions. Models were implemented using PyTorch 1.5, and experiments were conducted on a 64-bit Dell desktop with an NVIDIA GeForce RTX 2070, 32 GB RAM, and Intel (R) Core i9-9900 CPU. 

\subsection{Overall performance of the predictors in comparison}
Performance is compared between Persistence Model, Historical Average, AutoRegressive Integrated Moving Average (ARIMA), Feedforward Neural Network (FNN), LSTM (GeoTab), and LSTM (GeoTab+WU). All models are tested on same GeoTab stations with a missing data ratio up to 5.5\%, to predict the 24-hour surface temperature for the predetermined 27 testing days, given 48-hour previous observations.

\begin{itemize}
    \item \textbf{Persistence Model} is one of the simplest methods for predicting the future behavior of a time series. Persistence implies that future values of the time series are calculated on the assumption that conditions remain unchanged between “current” time and future time $t + T_H$ \cite{inman2013solar}.    
    \item \textbf{Historical Average} is calculated as the average value of the previous two days.
    \item \textbf{ARIMA} is a class of models that explain data using time series data on its past values and use linear regression to make predictions. Assuming that data has an autoregression relationship with its past values, the ARIMA model uses the dependent relationship between the current value and the past values within the time series. ARIMA is also a moving average model, where the model’s forecast depends linearly on its past values. To achieve the ARIMA model's best performance, we used the Auto ARIMA without tuning the required parameters of ARIMA. The Auto ARIMA model generates the optimal p, d, and q values suitable for the data set to provide better forecasting. 
    \item \textbf{FNN} is a multi-layer perceptron with additional hidden nodes between the input and the output layers. In this network, data moves in the only forward direction without any cycles or loops \cite{urso2018data}. The task of this research is to produce multi-horizon time series predictions, so it is essential to design an FNN with multiple outputs. Each neuron in the output layer focuses on the prediction of the considered variable at a different time step. The main issue with this architecture is that it does not take into account that the outputs are sequential (i.e., the same variable at different time steps). In fact, the model would act in the same way if the outputs were to predict different system variables at the same time step. 
    \item The \textbf{LSTM (GeoTab)} model was trained using only GeoTab data, and the \textbf{LSTM (GeoTab +WU)} model was trained using both GeoTab and WU data. The two LSTM models used same-size networks to predict on the testing dataset. 
\end{itemize}

\Table{table3} shows that our proposed method - LSTM (GeoTab+WU) model - achieves the best performance in RMSE. Historical Average achieves the second-best performance out of all considered models, especially better than LSTM (GeoTab). There is the possibility that the short-term availability of GeoTab dataset limited the capability of the LSTM architecture. All the considered models achieve a mean RMSE within the range of 3-4°C, indicating that using the previous 48 hours to predict the future 24 hours on fine-scale GeoTab stations is not an easy task.

\begin{table}[ht!]
\centering
\caption{Performance comparison}
\label{table3}
\begin{tabular}{l|l|l}
\hline
 & \textbf{RMSE (min/mean/max)} & \textbf{Bias (min/mean/max)} \\ \hline
\textbf{Persistence model} & 3.24 / 3.80 / 4.20 & 1.56 / 2.26 / 2.65 \\ \hline
\textbf{Historical average} & 3.36 / 3.71 / 4.03 & 0.4 / 0.49 / 0.61 \\ \hline
\textbf{ARIMA} & 3.29 / 3.98 / 4.56 & 0.34 / 0.92 / 1.52 \\ \hline
\textbf{FNN} & 3.59 / 3.86 / 4.13 & 0.67 / 0.95 / 1.2 \\ \hline
\textbf{LSTM (GeoTab)} & 3.49 / 3.77 / 4.02 & 0.22 / 0.55 / 1.1 \\ \hline
\textbf{LSTM (GeoTab + WU)} & 2.71 / 2.99 / 3.31 & 0.19 / 0.57 / 0.97 \\ \hline
\end{tabular}
\end{table}

\begin{figure}[ht!]
\centering
\includegraphics[width=0.8\textwidth]{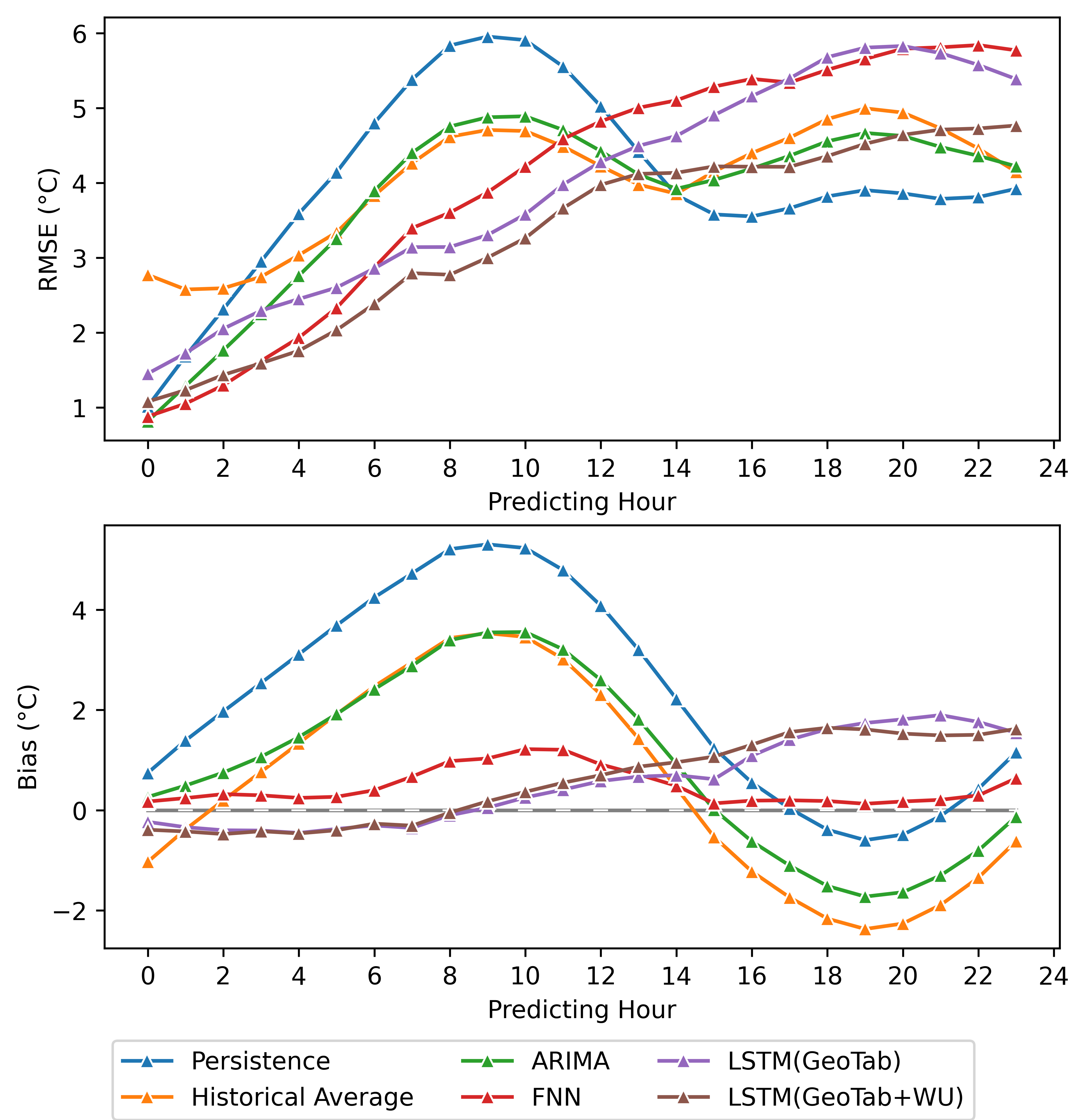}
\caption{
RMSE and Bias scores obtained with the six predictors. Performance computed on the test dataset.
}
\label{fig3}
\end{figure}

The RMSE and Bias scores for each predicting hour during testing are reported in \Fig{fig3}. The Persistence Model, Historical Average, and ARIMA showed similar sine wave patterns due to the diurnal change of surface temperature, where the largest overestimating error occurs at the 11th predicting hour (i.e., local noontime), and the largest underestimating error occurs at the 19th predicting hour (i.e., local 8 pm). On the contrary, the performance of the three neural networks - FNN, LSTM (GeoTab), and LSTM (GeoTab+WU) - decays with the predicting future hour. These three neural architectures are able to learn and predict the diurnal change successfully and seem to show a relatively similar predictive power, with the FNN predictor providing the most unsatisfactory performances. One possibility of the poor performance of FNN is that the FNN architecture explicitly used in this study is a multi-output network, with each output neuron irrelevant to other output neurons.

On the contrary, the LSTM models, including LSTM (GeoTab), and LSTM (GeoTab+WU), propagate the information through time remembering past inputs and reproducing the nonlinear function. The results show that training the LSTM predictors strongly mitigates this issue, supporting our hypothesis that this training method allows the proper information flow over subsequent time steps. The LSTM (GeoTab+WU) model showed a similar overall trend over the predicting hours to the LSTM (GeoTab) model, with progressively decreasing RMSE errors.

\subsection{Sensitivity of GeoTab missing data ratio}

\begin{figure}[!ht]
\centering
\includegraphics[width=0.8\textwidth]{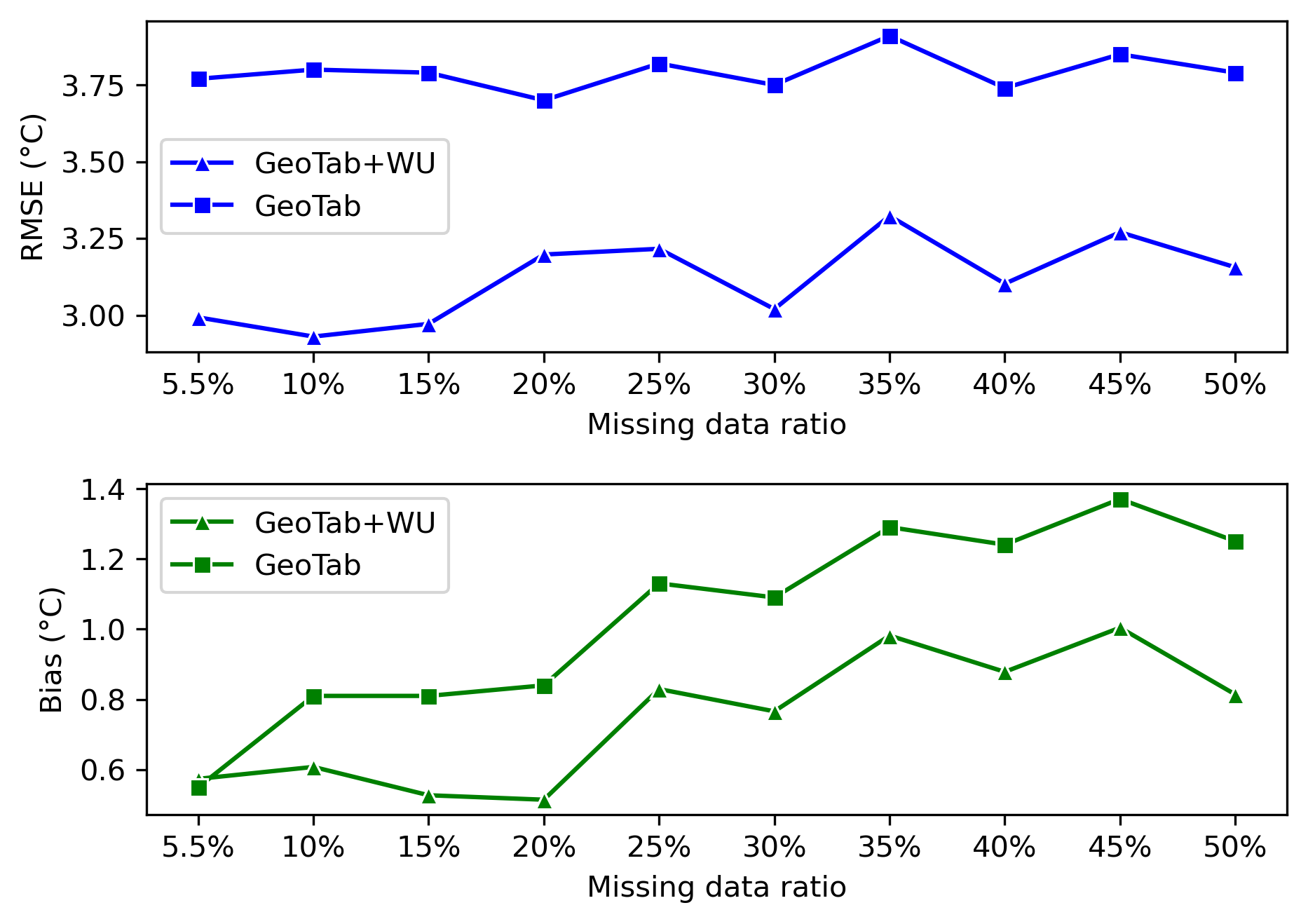}
\caption{
Mean RMSE and Bias errors for the same testing data generated by models trained using GeoTab stations with different missing data ratios. Errors are also shown for models trained using the GeoTab stations and WU historical data.
}
\label{fig4}
\end{figure}

To understand the impact of missing data on our proposed approach's predictability, a sensitivity experiment was conducted by training the model with stations having a missing data ratio from up to 5.5\% to 50\%, with 5\% as the increment (\Fig{fig4}). For the model trained by GeoTab only and the model trained by GeoTab+WU, integrating GeoTab stations with more missing data, the testing errors generally increase with noticeable fluctuations. The errors are primarily caused by the spatiotemporal interpolation while estimating the values for the inconsistent missing data. The fluctuating pattern might be the impact of cutting off stations using missing data ratio without considering the spatial continuity of stations. 

In addition, models trained by GeoTab showed less fluctuation than the ones trained by GeoTab +WU in RMSE errors, indicating that models are less sensitive to the increasing missing data ratio. On the contrary, models trained by GeoTab+WU showed more obvious fluctuation and larger increases in RMSE errors, indicating that models are more sensitive to the increasing missing data ratio. One possible reason is that the WU dataset is not interpolated and it represents real-world observations. By adding more GeoTab stations with a higher interpolation rate whose observations are smoothed, the model started to negotiate with both WU and IoT stations to learn the smoothed pattern, which lead to a larger fluctuation. Another possible reason for the different impacts to the two type of models is that GeoTab and WU have different value ranges of surface temperature and GeoTab show a larger spatiotemporal variability of surface temperature. When training the models with historical WU data and GeoTab, the models are impacted by the historical WU data more than GeoTab data, where WU had a longer period and a more constrained spatiotemporal variability. Involving GeoTab stations with more missing data downgraded the performance of the models trained by primarily WU. 

\subsection{Impact of adding historical WU in training}

To understand the impact of adding historical WU data to the training, a comparison experiment was conducted to examine whether the historical data can improve the local weather predictions. \Fig{fig4} showed the performance of models trained on GeoTab and GeoTab+WU stations. It's clear to see when the WU dataset is added to the training, the RMSEs are improved $\sim$20\% (mean value: 3.1 vs 3.8°C), and the overestimating bias is reduced for most of the experiments expect for the one trained using 5.5\% missing data. 

\begin{figure}[H]
\centering
\includegraphics[width=0.85\textwidth]{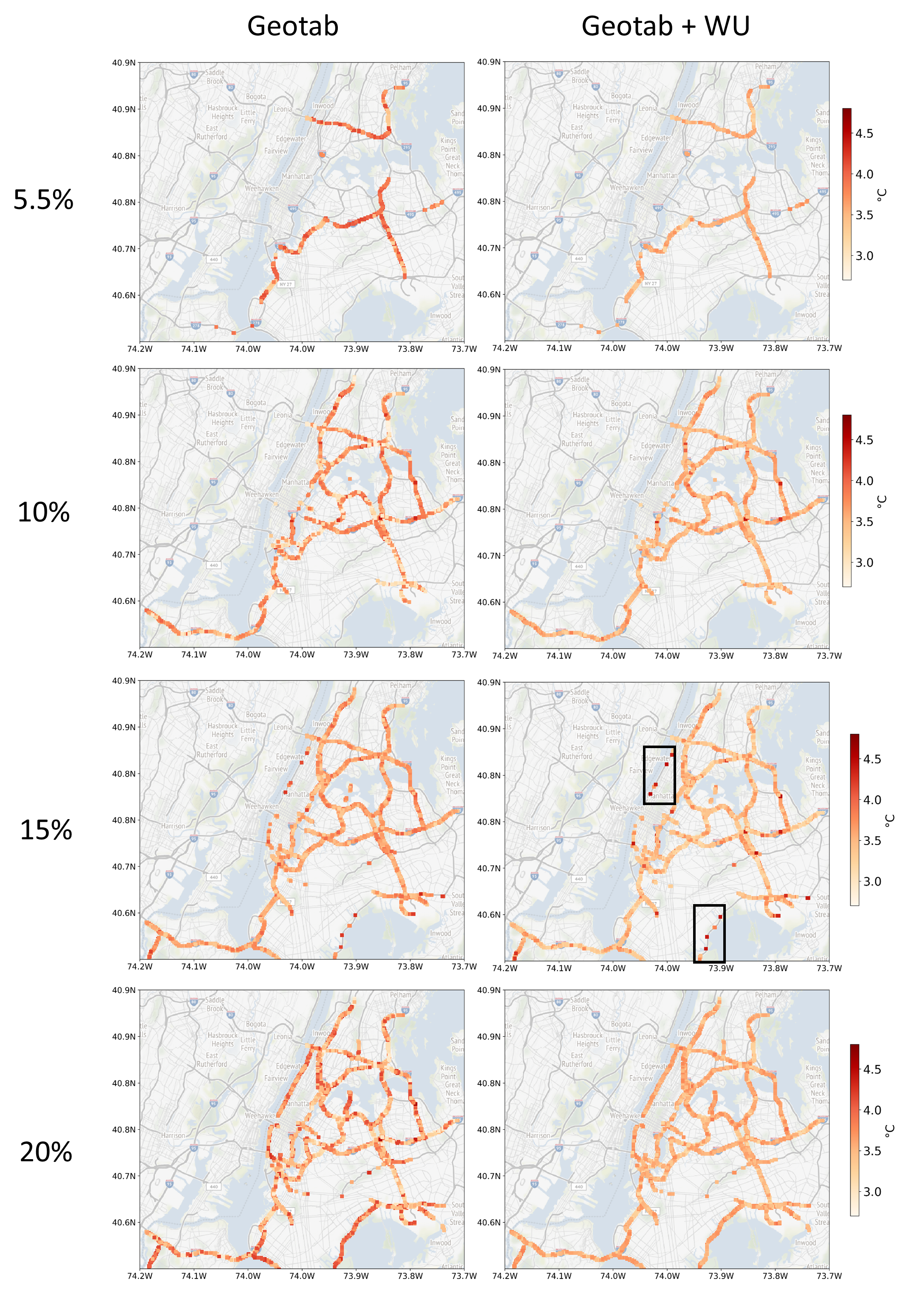}
\caption{
Spatial distributions of RMSE averaged for all testing days for each testing station. Comparison of models trained by GeoTab only and GeoTab + WU. 
}
\label{fig5}
\end{figure}

To investigate the spatial distribution of performance, another comparison experiment was conducted to train and test the LSTM model with GeoTab or GeoTab+WU on stations with different missing data ratio. \Fig{fig5} showed that LSTM (GeoTab) and LSTM (GeoTab+WU) have similar spatial patterns of mean RMSE, whereas adding historical WU data in training significantly reduced the RMSEs for most of the stations, except for the isolating stations. Note that there are a few stations along the river showed worse result for 15\% set when use Geotab +WU. This effect is due to the fact that these stations are located next to the river far from other stations and are very sparsely distributed in space. This problem is solved when we increase the GeoTab stations to 20\% because more GeoTab stations along the river in those regions are selected into training, and the network learned their patterns successfully.

\section{Discussions}

\subsection{Comparison with HRRR}
\begin{figure}[H]
\centering
\includegraphics[width=1.0\textwidth]{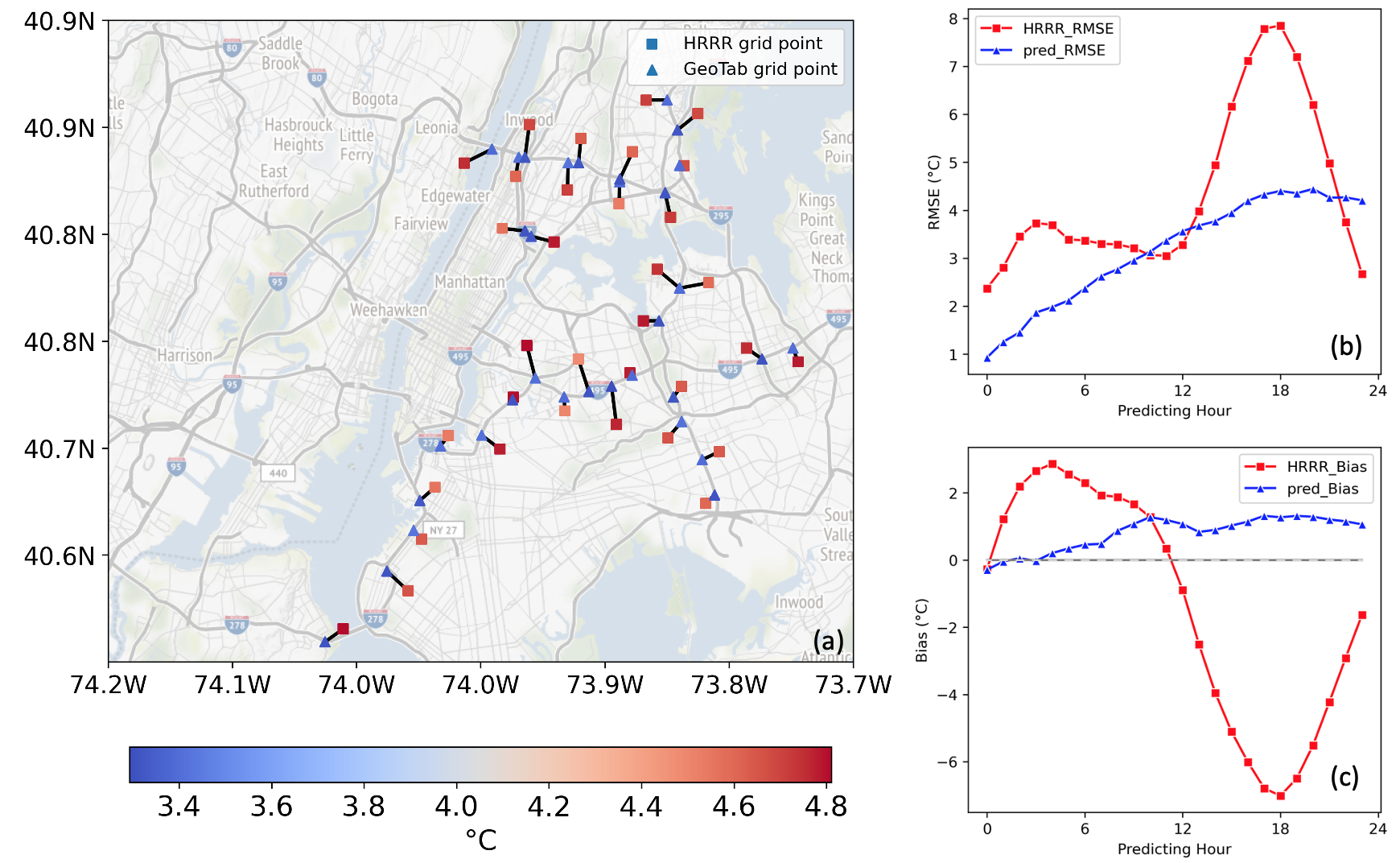}
\caption{
Comparing RMSEs between HRRR predictions and LSTM (GeoTab+WU) predictions at adjacent GeoTab stations. 
}
\label{fig6}
\end{figure}

To compare the air temperature predictions between the proposed framework and the numerical simulation, we also downloaded the High-Resolution Rapid Refresh (HRRR) predictions. The HRRR model is a 3km resolution, hourly updated atmospheric model. Radar data is assimilated in the HRRR every 15 min over a 1-h period. The dataset used in this study is downloaded from the University of Utah HRRR archive \cite{blaylock2017cloud}. \Fig{fig6}a showed the spatial locations of the HRRR grid points, and each HRRR grid point is compared with a GeoTab station within 2km. Each pair of HRRR and GeoTab grids are linked with a black line. The RMSEs are color coded in \Fig{fig6}a, with the RMSEs of HRRR predictions are visualized on each HRRR grid point, and the RMSEs of LSTM (GeoTab+WU) predictions are visualized on each GeoTab grid point. The RMSEs shown in this figure are the mean RMSE over all testing days. A remarkable observation is that LSTM (GeoTab+WU) predicted with a $\sim$30\% less RMSE than HRRR (mean RMSE: 3.64 vs. 4.61 °C). 

The performance is also decomposed into the 24 predicting hours in \Fig{fig6}b and \Fig{fig6}c. The HRRR predictions showed the diurnal change of predicting error in a sine wave pattern, and the amplitude of the sine wave during the local afternoon time is larger than the one in the local morning, partly due to the propagating error of the prediction (\Fig{fig6}b). The Bias values of HRRR predictions showed overestimations during the morning time and underestimations during the afternoons. The LSTM (GeoTab+WU) predictions showed an increasing RMSE over the 24 predicting hours, but the increase stopped at around Hour 18. In addition, LSTM (GeoTab+WU) predictions are generally overestimating the surface temperature, but the overestimation is generally lower or around 1 °C (\Fig{fig6}c). 

\subsection{Extreme cases}

\begin{figure}[H]
\centering
\includegraphics[width=0.85\textwidth]{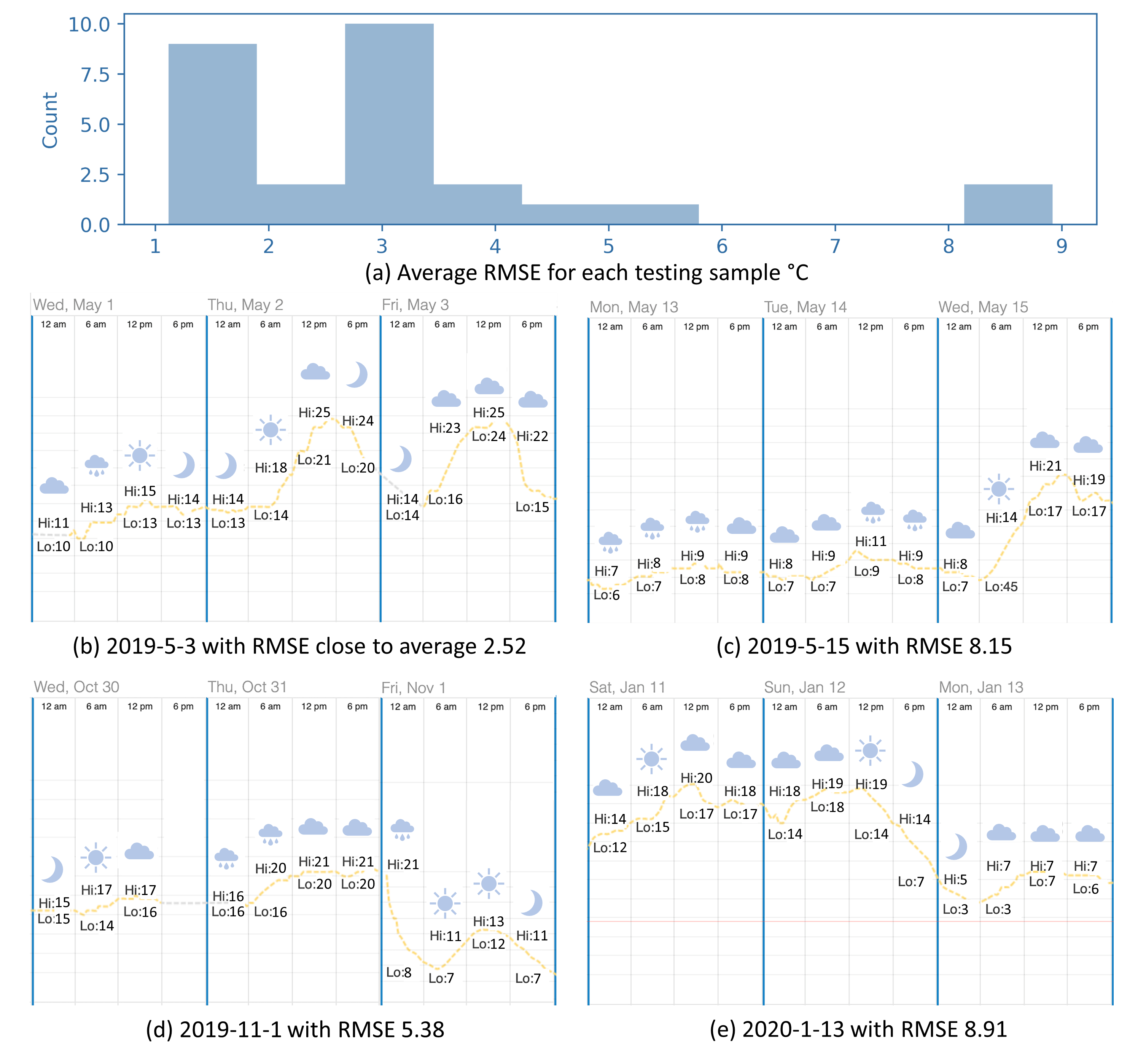}
\caption{
Histogram of RMSEs for each testing sample and demonstration of extreme weather cases. 
}
\label{fig7}
\end{figure}

We further investigated the model performance of LSTM (GeoTab+WU) over different testing days. For the 27 testing days, the overall average RMSE is 2.99. However, the RMSEs of some selected days are much higher than others. \Fig{fig7}a shows the histogram of average RMSE over 1650 stations for each selected day. The trained model performed well for most of days with an average RMSE under 4, and \Fig{fig7}b represents as one example. There are three selected days 2019-5-15, 2019-11-1, and 2020-1-13 that the model failed to predict well, and these days were confirmed to be rapidly changed weathers based on the records of weather stations (\Fig{fig7}c, \Fig{fig7}d, and \Fig{fig7}e). These three days have distinct temperature patterns from the previous two days. Note that the performance is highly related to the lengths of time series used as previous values and target values, i.e., 48 and 24 hours. Possible improvements can be made by integrating regional weather forecasts for long-term weather projections.

\section{Conclusions}

In this paper, we proposed a framework by integrating long-term historical in-situ observations and IoT observations to train a Long Short-Term Memory (LSTM) network for air temperature prediction within the city of New York. We compared the proposed framework with other time series prediction methods, specifically Persistence Model, Historical Average, AutoRegressive Integrated Moving Average (ARIMA), and Feedforward Neural Network (FNN). The LSTN network was trained in two differeny ways: 1) LSTM (GeoTab): using the IoT observations alone, and 2) LSTM (GeoTab+WU): using the IoT observation and the historical records from weather stations. Results showed that our proposed framework of integrating historical weather observations significantly improved the predictive performance of the LSTM network, and outputformed the other statistical and deep learning based time series prediction methods. By leveraging the historical air temperature data from in-situ observations, the LSTM model can be exposed to more historical patterns that might not be present in the IoT observations. Meanwhile, by using IoT observations, the spatial resolution of air temperature predictions is significantly improved. 

\bibliographystyle{plainnat}
\bibliography{mybibfile}

\end{document}